\documentclass[conference]{IEEEtran}

\IEEEoverridecommandlockouts

\usepackage{booktabs}

\usepackage{algcompatible}
\usepackage{algorithm}
\algblockdefx{FORP}{ENDFP}[1]
  {\textbf{for }#1 \textbf{do in parallel}}
  {\textbf{end for}}

\usepackage{tabularx}
\makeatletter
\newcommand{\multiline}[1]{%
  \begin{tabularx}{\dimexpr\linewidth-\ALG@thistlm}[t]{@{}X@{}}
    #1
  \end{tabularx}
}
\usepackage{multirow}
\makeatother

\usepackage{cite}
\usepackage{amsmath,amssymb,amsfonts}
\usepackage{graphicx}
\usepackage{textcomp}
\usepackage{xcolor}
\def\BibTeX{{\rm B\kern-.05em{\sc i\kern-.025em b}\kern-.08em
    T\kern-.1667em\lower.7ex\hbox{E}\kern-.125emX}}

\usepackage{xcolor, soul}

\usepackage{stfloats} 

\begin{document}

\title{AVARS - Alleviating Unexpected Urban Road Traffic Congestion using UAVs\\
\author{Jiaying Guo$^{1}$, Michael R. Jones$^{2}$, Soufiene Djahel$^{3}$, and Shen Wang$^{1}$
\thanks{$^{1}$Jiaying Guo and Shen Wang are with the School of Computer Science, University College Dublin, Ireland, E-mail: jiaying.guo@ucdconnect.ie, shen.wang@ucd.ie}%
\thanks{$^{2}$Michael R. Jones is with the Department of Computing and Mathematics, Manchester Metropolitan University, UK, E-mail: michael.jones@mmu.ac.uk}%
\thanks{$^{3}$Soufiene Djahel is with the Department of Computer Science, University of Huddersfield, UK, E-mail: s.djahel@hud.ac.uk}%
}
}

\maketitle

\begin{abstract}
Reducing unexpected urban traffic congestion caused by en-route events (e.g., road closures, car crashes, etc.) often requires fast and accurate reactions to choose the best-fit traffic signals. Traditional traffic light control systems, such as SCATS and SCOOT, are not efficient as their traffic data provided by induction loops has a low update frequency (i.e., longer than 1 minute). Moreover, the traffic light signal plans used by these systems are selected from a limited set of candidate plans pre-programmed prior to unexpected events' occurrence. Recent research demonstrates that camera-based traffic light systems controlled by deep reinforcement learning (DRL) algorithms are more effective in reducing traffic congestion, in which the cameras can provide high-frequency high-resolution traffic data. However, these systems are costly to deploy in big cities due to the excessive potential upgrades required to road infrastructure. In this paper, we argue that Unmanned Aerial Vehicles (UAVs) can play a crucial role in dealing with unexpected traffic congestion because UAVs with onboard cameras can be economically deployed when and where unexpected congestion occurs. Then, we propose a system called ``AVARS" that explores the potential of using UAVs to reduce unexpected urban traffic congestion using DRL-based traffic light signal control. This approach is validated on a widely used open-source traffic simulator with practical UAV settings, including its traffic monitoring ranges and battery lifetime. Our simulation results show that AVARS can effectively recover the unexpected traffic congestion in Dublin, Ireland, back to its original un-congested level within the typical battery life duration of a UAV.
\end{abstract}


\section{Introduction}
Many governments, policy makers, tech giants and SMEs are joining their efforts to support the vision towards achieving Net-Zero transport by 2050 (e.g., in the UK \& Ireland). This requires revolutionary solutions, supported by adequate policy changes, to a number of problems facing the wide adoption and/or deployment of cutting-edge technologies such as transport electrification, and the use of drones to offload ground transport, etc. Achieving the above objective also requires innovative solutions to better control traffic congestion, especially after the occurrence of unexpected events (e.g., car crashes and unplanned road work etc.) on the road. Unexpected (aka ``non-recurrent") urban road traffic congestion is caused by such en-route events. It is challenging to accurately predict the traffic impact of such events at any possible time and location due to the lack of such historical data. Therefore, an effective approach to reduce such congestion should react quickly after the event occurrence, rather than predicting the event impact in advance. This fast reaction is effective in reducing additional delays experienced by drivers and the resulting additional emissions. 

UAVs are renowned for their rapid deployment and efficiency in supporting emergency management (e.g., life-critical services \cite{bauer2021development}, situations where the communication infrastructure is destroyed, subject to cyber attacks or overloaded \cite{zhang2018fast, masroor2021efficient}). UAVs have also been used for road traffic monitoring on urban and highway roads \cite{liu2019vehicle}. However, the use of UAVs for emergency management in the transportation domain (e.g., reducing the impact of non-recurrent congestion), has not been sufficiently studied in the literature, thus we argue in this paper that UAVs have a great potential to be explored in this context. Additionally, although existing adaptive traffic light control systems (e.g., SCATS \cite{lowrie1990scats}) work well for the recurrent urban traffic, they cannot be effective for non-recurrent congestion due to their low-frequency and coarse-grained traffic monitoring detectors (i.e., induction loops collect traffic volumes in minutes). Recent DRL-based traffic light signal control can be promising \cite{wei2018intellilight}. However, the convergence of the DRL solution remains uncertain when applied in a fast-changing environment. Moreover, the practical deployment may need excessive hardware and software to upgrade existing systems (i.e., install cameras to all major road intersections and deploy DRL-based software on the road side units and central servers). This makes it unnecessary for non-recurrent congestion reduction which only happens very occasionally. To fill the above-mentioned gaps, we propose a system, dubbed AVARS, which is also open-sourced\footnote{https://github.com/Guojyjy/AVARS} and its main contributions are summarized as follows:

\begin{itemize}
    \item \textbf{AVARS uses UAVs to reduce non-recurrent congestion, rather than just monitoring normal traffic.} To the best of our knowledge, this is the first work that leverages UAVs to reduce non-recurrent congestion in urban traffic, while most others are for traffic monitoring, which we think is not practical as UAVs can only sustain for less than one hour compared to 24/7 deployed cameras in the fixed location. UAVs can be dispatched to congested intersections in a few minutes and provide high-frequency high-resolution data \cite{park2018usability}, which are well-suited features for recovering from non-recurrent road congestion within a short duration.
    \item \textbf{AVARS is a DRL-based system that contains many practical deployment considerations.} AVARS is carefully designed to achieve the minimum possible upgrade to the existing systems. This includes embedding low-cost wireless communication modules for traffic light controllers only (not vehicles). Thus, it avoids strong assumptions about having full or mix-autonomy traffic with vehicle-to-everything (V2X) communication infrastructures. AVARS also has a practical DRL model (i.e., newly designed state, action, and reward) as it can obtain a more stable convergence under fast-changing road traffic.
    \item \textbf{The effectiveness of AVARS is validated on a widely-used simulator with a realistic urban scenario.} The simulation study shows that AVARS can achieve the largest reduction in travel time, fuel consumption, and CO\textsubscript{2} emission, when unexpected congestion occurs in a city center area of Dublin, Ireland. Compared methods include SCATS \cite{lowrie1990scats} (no DRL, no UAV), and IntelliLight \cite{wei2018intellilight}(DRL but no UAV). Additionally, AVARS can recover non-recurrent congestion back to the original traffic within the 30-minute operation of UAVs, which is well within the normal UAV's battery life. If considering more realistic factors impacting battery duration, only 10-minute traffic signal control by AVARS is sufficient to mitigate congestion significantly.
\end{itemize}

\section{AVARS - System Description}

\subsection{Assumptions}

\textbf{Communications:} The communications required in our AVARS are only between a UAV and a traffic light controller (i.e., vehicles are not involved) that occur every second (i.e., to be responsive to en-route events). The communication technologies used could be any easy to implement ones such as 4/5G or IEEE 802.11p. In particular, the UAV send the ``action" to traffic lights to either switch to the next phase or keep the current phase. Conversely, the traffic light controller can send current traffic light phase back to the UAV as part of current observations (states). This is for the UAV to decide about the recommended action in the next time interval. We assume the communication is entirely reliable without any packet loss and delay.

\textbf{Computations:} Most computation loads are in the centralized Traffic Management Center (TMC). Firstly, TMC uses simulators to mimic possible road closures to collect data on the traffic impact. This is necessary for AVARS to select which intersections (i.e. likely the most badly influenced by the events) that UAVs must control after the occurrence of road events. Secondly, the pre-training of DRL models also happens in this simulation. Thirdly, after each assigned UAV completes its task, it brings back to TMC the realistic DRL trajectory data (i.e., state, action, reward, next state...) when UAV is operating AVARS. The trajectory data is also used for pre-training our DRL models. In addition to the computation loads at the TMC side, UAVs also need to do the following two computation tasks: extract the traffic information from their cameras; and control the traffic light controllers using a pre-trained DRL policy function with the real-time state information being the input.

\textbf{UAV Placement:} Upon the receipt of an intersection location (i.e., where traffic events occurred) from the TMC, it is assumed that a UAV can independently transit both safely and efficiently across the environment space. Upon arrival at the intersection location the UAV can maintain both its position and altitude, to deliver a visual representation of the intersection allowing for traffic information extraction.

\subsection{Design objectives}
We design AVARS to meet the following objectives:

\begin{itemize}
    \item \textbf{Be practical:} The system should have a practical deployment plan to meet the limitations of UAVs and DRL approaches. The cameras of UAVs have limited monitoring range. This implicates the maximum length of the road for UAV traffic monitoring should be around 220 meters \cite{park2018usability}. Moreover, the limited battery life of a UAV requires that the congestion should be effectively alleviated before the energy runs out (e.g., within 40 minutes since a UAV starts to operate \cite{hashemi2019new}). Additionally, the practical requirement of DRL refers to a stable convergence performance within reasonable amount of iterations. A stable DRL approach can potentially save computation resources and be effective in more general scenarios.
    \item \textbf{Be effective:} The system should be effective to recover the unexpectedly congested traffic within a limited time duration (i.e., UAV battery lifetime). The design of DRL models in terms of state, action, and reward is of vital importance to achieve this objective. The verification of this objective will be done by comparing the traffic evolution when no road closure occurs, as well as the reduction in travel time, fuel consumption, and CO\textsubscript{2} emissions.
\end{itemize}

\subsection{System components}
We clarify the responsibilities of each different AVARS component as follows:
\begin{itemize}
    \item \textbf{UAVs}: Each UAV is responsible for data collection (i.e., DRL state for decision making and DRL trajectories for pre-training) and sending actions (i.e., switching or not to the next traffic light phase) according to the learned DRL policy function for controlling a single intersection. 
    \item \textbf{Traffic lights}: Each traffic light only needs to receive the DRL action from a UAV to set its phase in the next time interval, and reports back to this UAV about the current phase as a part of DRL state.
    \item \textbf{TMC}: Based on numerous simulation results, TMC needs to select a number of key intersections for UAVs to control when unexpected events occur. TMC also needs to train the proposed DRL model using DRL trajectories collected after UAVs complete their AVARS field tasks.
\end{itemize}
Unlike most of existing works in the literature \cite{wei2019survey}, our AVARS does not rely on any vehicle equipped with advanced technologies such as autonomous or connected vehicles. This saves a large potential cost to upgrade existing infrastructure and removes strong assumptions on collecting vehicle driving information in recent DRL-based traffic light signal control approaches \cite{wang2021adaptive}.

\subsection{System flow}
\begin{figure}[!ht]
    \centering
    \includegraphics[width=\linewidth]{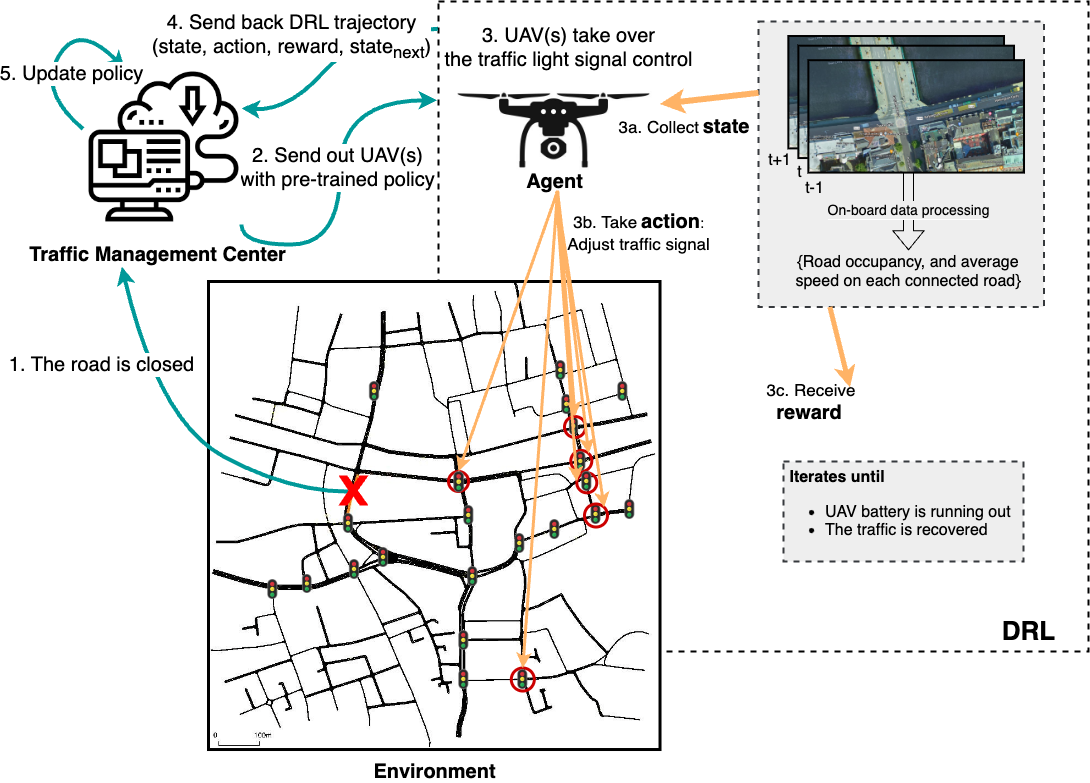}
    \caption{Illustration of AVARS system flow. The traffic environment is a subnet of Dublin city center road network. The highlighted road is the closed road causing unexpected congestion in the surrounding area. UAVs of AVARS control the six red-circled intersections in this scenario, which are influenced most by the road closure.}
    \label{fig:system}
\end{figure}

Suppose a road segment within an urban road network is suddenly closed, due to an incident, leading to queues of vehicles, reduced traffic speed and several affected vehicles needing to reroute to bypass the closed road. This will migrate the traffic demand to several other signalized intersections, which causes unexpected traffic congestion. To efficiently handle such scenarios, AVARS operates as follows (also shown in Fig.\ref{fig:system}):

\begin{enumerate}
    
    \item Once a road is closed, the TMC is informed through existing traffic monitoring technologies.
    \item The TMC suggests several signalized intersections for UAVs to control (six intersections highlighted by red circles in the environment of Fig.\ref{fig:system}). Each UAV is assigned to a designated intersection. For each selected intersection, all of its directly connected roads should be covered by the monitoring range of the UAV camera. 
    \item When UAVs arrive at their assigned intersections, each UAV starts: a) collecting the traffic information, then converting them to the ``DRL state"; b) sending the ``DRL action" to adjust the traffic signal according to the pre-trained DRL policy function; and c) calculating the ``DRL reward" based on real-time traffic information. These 3 steps iterate every second until the preset operation duration is reached or the UAV enters a low battery state, both triggering a UAV to return to the TMC. Each UAV returns to TMC with the collected DRL trajectory (state, action, reward, new state) acquired during this operation interval.
    \item The DRL policy function is updated using newly collected DRL trajectory data on the TMC cloud servers.
\end{enumerate}

\subsection{DRL models}
\label{sec:sys_drl}
\begin{itemize}
    \item State: For a specific intersection that is controlled by a UAV, the state involves the current signal phase of the traffic lights, the occupancy (i.e., the ratio of the total length of vehicles on a specific road to the length of that road) and average vehicle speed on each of the roads directly linked to this intersection. This state information is collected from the UAV's sensors (e.g., camera) and extracted after onboard data processing \cite{khan2018unmanned}.
    \item Action: The action that each UAV sends to its controlled traffic light is a binary set, which defines that ``1" represents switching to the next phase and ``0" means to keep the current phase unchanged for the next time interval. 
    \item Reward: Given the DRL state and action at the intersection $r_j$, as described in Eq.\ref{eq:reward}, the reward is calculated as the negative value of maximum road occupancy among all incoming roads $I$ connected to the intersection $j$. The reward aims to evacuate traffic on the road that has the highest occupancy as soon as possible.
    \begin{equation}
        r_j = -max(O_i), i \in I
        \label{eq:reward}
    \end{equation}
    The road occupancy $O_i$ for each incoming road is also involved in the UAV state.
\end{itemize}

To get a stable convergence with a moderate number of iterations, we do not recommend any multi-agent DRL algorithm for AVARS so far \cite{wei2019survey}. Algorithm 1 shows one completed iteration of AVARS training process to obtain the DRL policy function taken by UAVs to control traffic lights, in which the DRL algorithm can be either Deep Q-learning (DQN) or Proximal Policy Optimization (PPO) \cite{schulman2017proximal}. DQN is commonly used in existing traffic light signal control systems \cite{wei2018intellilight, wang2021adaptive}, having advantages in discrete action space, i.e., a binary set of traffic signal switching in AVARS. The off-policy method is efficient to update the finite Q values on a one-step temporal difference for a higher reward. However, it is also volatile in a fast-changing environment, especially with complex traffic information, which aggravates training convergence. Unlike DQN, PPO can reduce the large variance of policy updates and facilitate training convergence. PPO can also improve sample efficiency by multiple epochs of mini-batch updates. 

\begin{figure}[!ht]
  \label{alg:train}
  \renewcommand{\algorithmicrequire}{\textbf{Input:}}
  \renewcommand{\algorithmicensure}{\textbf{Output:}}
  \
  \begin{algorithm}[H]
    \caption{One DRL training iteration of AVARS}
    \begin{algorithmic}[1]
        \REQUIRE A set of intersections $E$ to control
        \ENSURE Policy function parameter $\theta$
        \STATE Initialize the urban road environment in $T$ time horizons constrained by UAV battery lifetime
        \STATE Initialize  Q-network $Q(\theta)$ if DQN or policy network $\pi(\theta)$ if PPO
            \FORP{each $E$}
                \STATE UAV arrives at the designated intersection
                \FOR{$t \leq T$}
                    \STATE Get observation $s_t$
                    \State \multiline{%
                    Execute action $a_t$ selected through $Q(s_t, a; \theta)$ or $\pi(a|s_t; \theta)$, $a \in \{0, 1\}$ and transit to $s_{t+1}$}
                    \State \multiline{%
                    Calculate reward $r_t$ using Eq.\ref{eq:reward}}
                    \State \multiline{%
                    Collect trajectory $\tau = (s_t, a_t, r_t, s_{t+1})$ and extend $B$ with $\tau$ }
                \ENDFOR
        \ENDFP
        \State \multiline{%
        Sample a random batch from $B$ to update the policy parameter $\theta$}
    \end{algorithmic}
  \end{algorithm}
\end{figure}

\subsection{Summary}
We summarize how the design of AVARS achieves our first objective - \textit{be practical}. Firstly, AVARS avoids large-scale upgrades of hundreds of traffic light controllers by quickly assigning UAVs to very few intersections that are most impacted by the en-route events. Secondly, AVARS uses UAVs' sensors to obtain real-time traffic information, which is much more flexible than embedding induction loops on the road. Thirdly, AVARS does not consider vehicle-related upgrades (e.g., V2X communication capabilities) and requires minimum upgrades of the traffic lights (e.g., 5G or IEEE 802.11p communication technology) to communicate with UAVs instead of the costly wired cable communications used in today's adaptive traffic light systems. Last but not least, our DRL model is simple enough to calculate. Additionally, the heavy-weighted computation for training DRL only occurs in the centralized TMC, which is practical to implement using existing cloud-based services, rather than distributing these computation loads to multiple roadside units or resource-constrained UAVs. 

\section{Evaluation Results and Analysis}

\subsection{Unexpected urban traffic congestion scenarios}
The implementation of AVARS is based on FLOW \footnote{https://flow-project.github.io}, which is a framework integrating RLlib (DRL library) and SUMO (traffic simulator).
The testing map is a subnet of Dublin city center road network around the River Liffey, covering approximately 1 square kilometer, as shown in Fig.\ref{fig:system}. The scenario is extracted from the open data in \cite{gueriau2020quantifying} to simulate the real-world traffic in Dublin city, and the traffic generation lasts 45 minutes with 1168 vehicles in our experiments. We close a road in the center of the testing scenarios for 30 minutes from the 10th minute of the simulation, at which point the vehicles that would have entered the closed road are required to reroute. The radius of the area managed by the TMC at Dublin city center is 5km, given the maximum flying speed of 80 km/h \cite{khan2018unmanned}, a UAV is capable of direct flight to any intersection within the control region in 3.75 minutes. However, the direct flight may be limited by the urban environment, hence we assume the start time of UAV control after road closure is 5 minutes.

\subsection{Compared scenarios}
\begin{itemize}
    \item \textbf{Original}: This scenario contains the regular traffic for a given urban region. It is a reference to verify if the unexpected congestion is recovered. The traffic light signals are static in the scenario, operating on a fixed cycle plan. The duration of a signal plan ranges from 90 seconds to 113 seconds for the given heterogeneous intersections. Green light phases last from 27 seconds to 50 seconds. Every time the green light switches, a 3-second yellow is followed to clear the intersection, and some signal plans combine a 5-second all-red light after the yellow light. 
    \item \textbf{Congestion}: This scenario contains unexpected congestion due to one road closure for the same urban region. The traffic light signals are static, the same as the settings in the Original scenario.
    \item \textbf{SCATS \cite{lowrie1990scats}}: This scenario contains the simulation of the existing adaptive traffic light controller, SCATS \cite{lowrie1990scats}, which neither use DRL nor UAVs. The embedded induction loops detect the duration of vehicles passing through the intersection during the green light phase, which can estimate the effective green time. The sum of the ratio of effective green time for each green light phase is the degree of saturation of one signal plan. SCATS provides multiple pre-defined signal plans and selects the signal plan with the minimum degree of saturation \cite{wei2019survey}. 
    \item \textbf{IntelliLight \cite{wei2018intellilight}}: This scenario contains the DRL-based traffic light signal control model using DQN but not using UAVs. Compared with AVARS, more complicated traffic metrics, such as queue length, number of vehicles, and waiting time of vehicles, are included in state and reward design. Besides, the state is also expanded with vehicles' positions and traffic light phases collected from the image representation. IntelliLight controls the signal switching to reduce queue length and travel delays.

\end{itemize}

\subsection{Result analysis}
\begin{figure}[htbp]
    \centering
    \includegraphics[width=\linewidth]{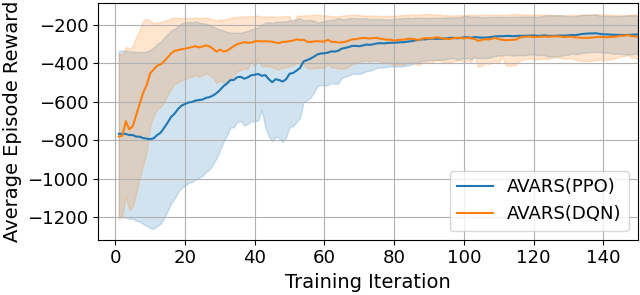}
    \caption{Evolution of the average episode reward during the training process of AVARS, computed over 18 parallel episodes at each iteration. As for AVARS using PPO, the reward ranges from approx. -766 to -250, and the standard deviation decreases from about 430 to 99.}
    \label{fig:AVARS_reward}
\end{figure}

\begin{figure}[htbp]
    \centering
    \includegraphics[width=\linewidth]{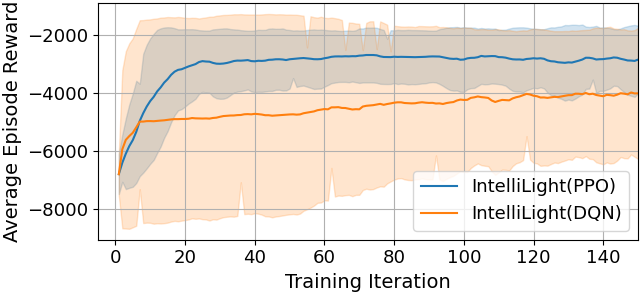}
    \caption{Evolution of the average episode reward during the training process of IntelliLight, computed over 18 parallel episodes at each iteration. The reward of IntelliLight(DQN) ranges from approx. -6800 to -4050, and the standard deviation decreases from about 3400 to 2200.}
    \label{fig:IntelliLight_reward}
\end{figure}

\textbf{Is DRL training of AVARS more stable to converge?}
As shown in Fig.\ref{fig:AVARS_reward}, after 150 iterations, the training of AVARS(PPO) terminates at a higher average episode reward and smaller standard deviation (approx. -250 and 99) than the initial stage (approx. -766 and 430), which are calculated over 18 parallel episodes for each iteration. It also demonstrates the continuous positive updates of the trained policy. The DQN training of AVARS gets similar results, while the policy update of DQN is much faster, resulting in earlier convergence. Instead, the reward improvement of IntelliLight shown in Fig.\ref{fig:IntelliLight_reward} varies widely across different DRL algorithms. PPO can get a better policy with a higher average episode reward (about -2860) after training compared to the DQN reward (about -4050). However, the standard deviation of either PPO or DQN during training is large, probably due to the complex design of states and rewards to evaluate traffic. Meanwhile, the complex design of IntelliLight(DQN) results in the training time being approximately three times longer than IntelliLight(PPO). Hence, the benefit of AVARS in terms of states and reward design can also be demonstrated by the above experiments on AVARS and IntelliLight using DQN or PPO. AVARS, with the concise state and reward design, can take advantage of DQN or PPO to get an expected policy to recover unexpectedly congested traffic, which is in line with the second system design objective - \textit{effectiveness}. Additionally, we would simplify the following experiments with only AVARS(PPO) and IntelliLight(DQN) because PPO is more stable than DQN in traffic improvement based on IntelliLight, while DQN is the default DRL algorithm for IntelliLight.

\begin{figure}[htbp]
    \centering
    \includegraphics[width=\linewidth]{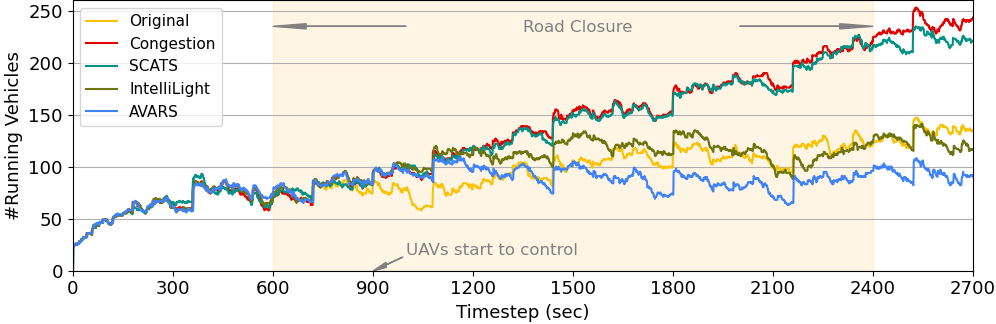}
    \caption{Evolution of the number of running vehicles in the chosen urban region over the simulated timesteps of an episode under AVARS and other compared scenarios. Road closure starts from the 600th timesteps and lasts 30 mins. UAVs used in AVARS start to control traffic light signals from the 900th timesteps.}
    \label{fig:compared_runVeh}
\end{figure}

\textbf{Is AVARS effective?}
Vehicles generated for the 45 minutes (i.e., 2700 timesteps), so the number of vehicles running in the simulated region tends to gradually increase during this time interval, as illustrated in the Original scenario of Fig.\ref{fig:compared_runVeh}. The road closure (the Congestion scenario) has resulted in a sharp increase in running vehicles, up to more than 250. However, after UAVs effectively control the selected intersections for approximately 10 minutes, the number of running vehicles decreases to a level lower than the Original scenario at the same timestep. The largest difference between the Original scenario and our proposed method is 61 vehicles around the 2500th timestep. SCATS cannot evidently mitigate the congested traffic, and in this scenario, the number of running vehicles shows only minor deviations from the Congestion scenario. The results of IntelliLight are closely aligned to that of the Original scenario. Hence, Fig.\ref{fig:compared_runVeh} illustrates that AVARS can alleviate the unexpected congestion.

Furthermore, we analyze the main traffic statistics including average travel time, fuel consumption, and CO\textsubscript{2} emissions (i.e., we refer to these metrics as ``main traffic metrics" in the following descriptions), as shown in Table.\ref{tab:statistics}. The results are the average of 10 episodes that terminated after all vehicles completed the trip. Our proposed system AVARS produces the best results. Even compared to the Original scenario, where no road closure is in place, all traffic metrics are significantly reduced. By contrast, the negative impact of one critical road closure is evident, with the value of these main traffic metrics more than doubling up to 126\%. Congestion causes massive delays, leading to longer travel times. 
 
All traffic light signal control methods, SCATS, IntelliLight, and AVARS, can alleviate the unexpected traffic congestion. Although SCATS reduces about 30\% on these main traffic metrics over the Congestion scenario, the results are still inferior to that of the Original scenario. The results show that SCATS is not efficient in alleviating unexpected traffic congestion. In comparison, both DRL-based methods achieve a substantial improvement in average travel time, fuel consumption, and CO\textsubscript{2} emissions, while our system AVARS achieves the largest reduction, 67\%, 72\%, and 72\%, respectively.
\begin{table}[htbp]
    \centering
    \caption{Main traffic statistics. Except for the Original scenario, there is a road closure under all other scenarios. The achieved traffic improvements of SCATS, IntelliLight, and AVARS are based on the Congestion scenario.}
    \begin{tabular}{m{1.5cm}<{\centering}|m{1.8cm}<{\centering}|m{1.8cm}<{\centering}|m{1.8cm}<{\centering}}
    \toprule
    & \textbf{Average travel time (s)} & \textbf{Fuel consumption (l/100km)} & \textbf{CO\textsubscript{2} emissions (g/km)} \\
    \midrule
    \multirow{1}{1.5cm}{\textbf{Original}} & 275.78 & 20.14 & 468.62 \\
    \midrule
    \multirow{2}{1.5cm}{\textbf{Congestion}} & 597.31 & 45.45 & 1057.20 \\
    & - & - & - \\
    \midrule
    \multirow{2}{1.5cm}{\textbf{SCATS}} & 429.61 & 31.69 & 737.10 \\
    & -28.08\% & -30.28\% & -30.28\% \\
    \midrule
    \multirow{2}{1.5cm}{\textbf{IntelliLight}} & 232.67 & 14.60 & 339.55 \\
    & -61.05\% & -67.88\% & -67.88\% \\
    \midrule
    \multirow{2}{1.5cm}{\textbf{AVARS}} & 195.42 & 12.31 & 286.29 \\
    & \textbf{-67.28\%} & \textbf{-72.92\%} & \textbf{-72.92\%} \\
    \toprule
    \end{tabular}
    \label{tab:statistics}
\end{table}

The travel time statistics for all vehicles also demonstrates the benefit of our system for the vehicles influenced by the unexpected congestion, as shown in Table.\ref{tab:traveltime_dis}. These vehicles have experienced more prolonged travel times due to the congestion. Except for the shortest average travel time, the smallest standard deviation of AVARS demonstrates a general reduction in vehicle congestion delays. More than half of vehicles can finish their trips in about 200 seconds, which is better than the Original scenario. In addition, the longest travel time is decreased from 4599 seconds in the Congestion scenario to 554 seconds. Vehicles can reach their destination earlier, even when travelling longer distances due to rerouting.

\begin{table}[htbp]
    \centering
    \caption{Travel time statistics (average, 25\% percentile, median, 75\% percentile, minimum, maximum, and standard deviation) for all vehicles across all compared scenarios.}
    \begin{tabular}{m{0.6cm}<{\centering}|m{1.0cm}<{\centering}|m{1.3cm}<{\centering}|m{0.8cm}<{\centering}|m{1.3cm}<{\centering}|m{0.8cm}<{\centering}}
    \toprule
    & \textbf{Original} & \textbf{Congestion} & \textbf{SCATS} & \textbf{IntelliLight} & \textbf{AVARS} \\
    \midrule
    \textbf{Avg} & 275.78 & 597.31 & 429.61 & 232.67 & 195.42 \\
    \midrule
    \textbf{Std} & 208.83 & 965.67 & 614.66 & 148.08 & 108.86 \\
    \midrule
    \textbf{Min} & 2.60 & 2.50 & 1.00 & 2.60 & 1.00 \\
    \midrule
    \textbf{1st Qtr} & 127.40 & 129.52 & 119.00 & 119.53 & 109.55 \\
    \midrule
    \textbf{Med} & 230.70 & 215.90 & 209.00 & 206.70 & 193.35 \\
    \midrule
    \textbf{3rd Qtr} & 387.78 & 342.48 & 333.00 & 304.85 & 261.00 \\
    \midrule
    \textbf{Max} & 1365.20 & 4599.10 & 3059.00 & 743.00 & 553.70 \\
    \toprule
    \end{tabular}
    \label{tab:traveltime_dis}
\end{table}

\begin{figure}[htbp]
    \centering
    \includegraphics[width=\linewidth]{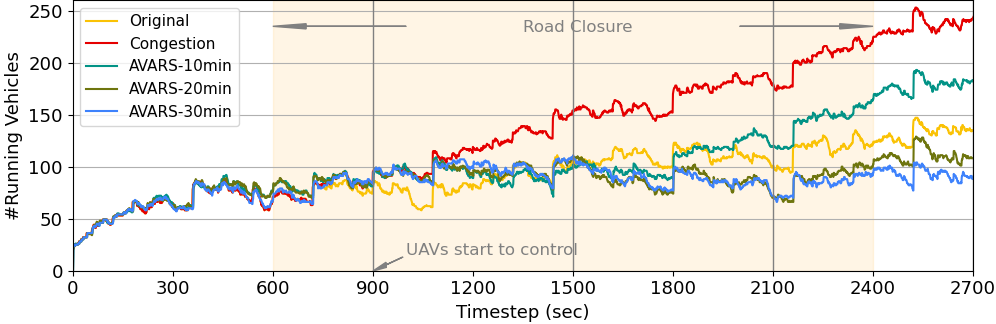}
    \caption{Evolution of the number of running vehicles in the chosen urban region over the simulated timesteps of an episode under different UAV operation durations of AVARS compared with the Original and Congestion scenarios. UAVs used in AVARS start to control traffic light signals from the 900th timesteps and terminate at the 1500th timestep (10 minutes), the 2100th timestep (20 minutes), and the 2700th timesteps (30 minutes), respectively.}
    \label{fig:runVeh_UAVtime}
\end{figure}

\textbf{How fast can AVARS alleviate the congestion?} 
Time periods of 10, 20 and 30 minutes are selected as operational AVARS durations, all shorter than 40 minutes, which represent the typical UAV battery lifetime \cite{hashemi2019new}. Fig.\ref{fig:runVeh_UAVtime} suggests that although the longer the UAV operates, the better traffic improvement can be achieved, just 10 to 20 minutes of UAV operation in AVARS should be sufficient to recover the congestion back to its original state.

\section{Conclusion and Future Work}

This paper proposes AVARS: a DRL-based traffic light control system using UAVs to reduce unexpected urban road traffic congestion. AVARS takes advantage of UAVs' rapid deployment, and rich traffic monitoring capabilities to timely react to these events. AVARS also supports state-of-the-art DRL methods to deliver an efficient control of traffic light signals, without the need for the costly upgrade of traffic light controllers. We demonstrated that AVARS can effectively alleviate congestion and achieve notable reductions in travel time, fuel consumption, and CO\textsubscript{2} emissions in an urban scenario. Future works may include designing an extended version of AVARS in which multiple UAVs cooperate using multi-agent DRL while addressing its high complexity of cooperation and convergence issues.

\bibliographystyle{IEEEtran}
\bibliography{main.bbl}

\begin{thebibliography}{10}

\bibitem{bauer2021development}
Jan Bauer, Dieter Moormann, Reinhard Strametz, and David~A Groneberg.
\newblock {Development of unmanned aerial vehicle (UAV) networks delivering
  early defibrillation for out-of-hospital cardiac arrests (OHCA) in areas
  lacking timely access to emergency medical services (EMS) in Germany: a
  comparative economic study}.
\newblock {\em BMJ open}, 11(1):e043791, 2021.

\bibitem{zhang2018fast}
Xiao Zhang and Lingjie Duan.
\newblock {Fast deployment of UAV networks for optimal wireless coverage}.
\newblock {\em IEEE Transactions on Mobile Computing}, 18(3):588--601, 2018.

\bibitem{masroor2021efficient}
Rooha Masroor, Muhammad Naeem, and Waleed Ejaz.
\newblock {Efficient deployment of UAVs for disaster management: a
  multi-criterion optimization approach}.
\newblock {\em Computer Communications}, 177:185--194, 2021.

\bibitem{liu2019vehicle}
Shaohua Liu, Suqin Wang, Wenhao Shi, Haibo Liu, Zhaoxin Li, and Tianlu Mao.
\newblock {Vehicle tracking by detection in UAV aerial video}.
\newblock {\em Science China Information Sciences}, 62(2):1--3, 2019.

\bibitem{lowrie1990scats}
PR~Lowrie.
\newblock {SCATS, sydney co-ordinated adaptive traffic system: A traffic
  responsive method of controlling urban traffic}.
\newblock 1990.

\bibitem{wei2018intellilight}
Hua Wei, Guanjie Zheng, Huaxiu Yao, and Zhenhui Li.
\newblock Intellilight: A reinforcement learning approach for intelligent
  traffic light control.
\newblock In {\em Proceedings of the 24th ACM SIGKDD International Conference
  on Knowledge Discovery \& Data Mining}, pages 2496--2505, 2018.

\bibitem{park2018usability}
Keunhyun Park and Reid Ewing.
\newblock {The usability of Unmanned Aerial Vehicles (UAVs) for pedestrian
  observation}.
\newblock {\em Journal of Planning Education and Research}, page
  0739456X18805154, 2018.

\bibitem{hashemi2019new}
Seyed~Reza Hashemi, Roja Esmaeeli, Haniph Aliniagerdroudbari, Muapper Alhadri,
  Hammad Alshammari, Ajay Mahajan, and Siamak Farhad.
\newblock New intelligent battery management system for drones.
\newblock In {\em ASME international mechanical engineering congress and
  exposition}, volume 59438, page V006T06A028. American Society of Mechanical
  Engineers, 2019.

\bibitem{wei2019survey}
Hua Wei, Guanjie Zheng, Vikash Gayah, and Zhenhui Li.
\newblock A survey on traffic signal control methods.
\newblock {\em arXiv preprint arXiv:1904.08117}, 2019.

\bibitem{wang2021adaptive}
Tong Wang, Jiahua Cao, and Azhar Hussain.
\newblock Adaptive traffic signal control for large-scale scenario with
  cooperative group-based multi-agent reinforcement learning.
\newblock {\em Transportation research part C: emerging technologies},
  125:103046, 2021.

\bibitem{khan2018unmanned}
Muhammad~Arsalan Khan, Wim Ectors, Tom Bellemans, Davy Janssens, and Geert
  Wets.
\newblock Unmanned aerial vehicle-based traffic analysis: A case study for
  shockwave identification and flow parameters estimation at signalized
  intersections.
\newblock {\em Remote Sensing}, 10(3):458, 2018.

\bibitem{schulman2017proximal}
John Schulman, Filip Wolski, Prafulla Dhariwal, Alec Radford, and Oleg Klimov.
\newblock Proximal policy optimization algorithms.
\newblock {\em arXiv preprint arXiv:1707.06347}, 2017.

\bibitem{gueriau2020quantifying}
Maxime Gu{\'e}riau and Ivana Dusparic.
\newblock Quantifying the impact of connected and autonomous vehicles on
  traffic efficiency and safety in mixed traffic.
\newblock In {\em 2020 IEEE 23rd International Conference on Intelligent
  Transportation Systems (ITSC)}, pages 1--8. IEEE, 2020.

\end{thebibliography}

\vspace{12pt}

\end{document}